\documentclass[12pt,twoside,a4paper,reqno]{amsart}
\usepackage{bulletinUPB}
\usepackage{graphics}
\usepackage{amssymb}
\usepackage{eucal}
\usepackage{amsmath}
\usepackage{hyperref}
\usepackage{graphicx}
\usepackage{float}
\usepackage{algorithm}
\usepackage{algpseudocode}
\usepackage{listings}
\usepackage{caption}

\info{C}{88}{2}{2026}
\title{Multi-modal video data-pipelines for machine learning with minimal human supervision}

\author{P\^{i}rvu Mihai-Cristian}
\address{$^1$PhD student, Institute of Mathematics of the Romanian Academy "Simion Stoilow", e-mail: {\tt mihaicristianpirvu@gmail.com}}

\author{Marius Leordeanu}
\address{$^2$Professor, Faculty of Automatic Control and Computer Science, National University of Science and Technology POLITEHNICA, Romania, e-mail: {\tt leordeanu@gmail.com}}

\begin{document}
\pagestyle{headings}
\maketitle

\begin{abstract}
Traditionally, Machine Learning models have been unimodal (i.e. $RGB \rightarrow semantic\ segmentation $ or $text \rightarrow sentiment\ class$), yet the real-world is inherently multi-modal. Capturing and correlating these modalities from raw video without manual annotation presents a significant engineering challenge. To address this, we introduce the Video Representations Extractor (VRE), an open-source, modular and configurable data pipeline designed to generate multi-modal datasets with minimal human supervision. By leveraging pre-trained neural networks and procedurally generated combinations of them, VRE can orchestrate the computation of many distinct modalities which can augment a video dataset with a much richer scene understanding. We describe the engineering architecture behind VRE, detailing strategies for scalability, including multi-GPU batching and real-time streaming. We validate our approach by highlighting the results of \cite{mihai2025probabilistic}, which successfully extends the Dronescapes dataset \cite{marcu2023self} with up to 13 new modalities and 16 new UAV videos using our tool. Finally, we showcase the framework's capability to deploy real-time inference models for semantic segmentation and depth estimation on commodity hardware. While we mostly focus on UAV datasets, the same procedure can be applied to any video domain, like autonomous driving or indoor robotics.
\end{abstract}


\section{Introduction and related work}

\begin{figure}[h]
    \includegraphics[width=1\linewidth]{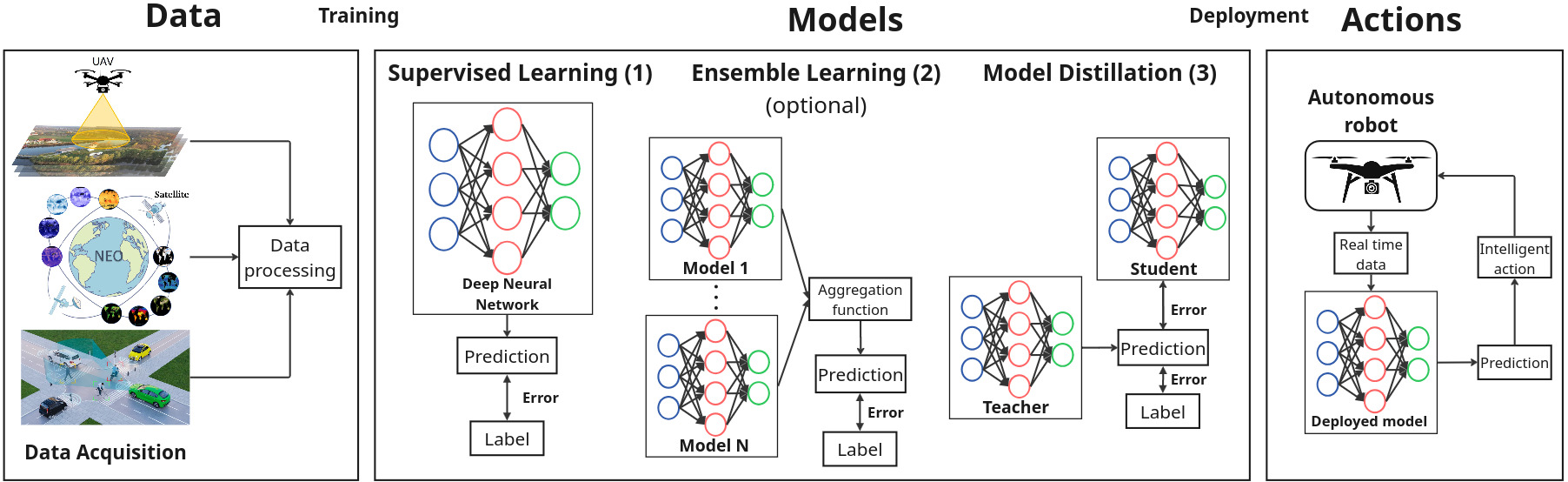}
    \caption{High-level overview of an end-to-end Machine Learning system: from raw data and data processing, to training and optimizing models and lastly, deploying it to interact and control a real hardware autonomously with intelligent actions.}
    \label{fig:main-figure}
\end{figure}

Machine Learning systems typically consist of three components: data (1), models and algorithms (2), and predictions and actions (3), as presented in Figure \ref{fig:main-figure}. Research has primarily focused on models tested on fixed benchmarks. This is practical: it allows direct, unbiased comparison against existing work in a controlled setting while reusing existing data processing methods. The approach has worked very well and has driven the progress of the field with results such as the AlexNet \cite{krizhevsky2012imagenet} classification network on the ImageNet dataset, the Transformer network \cite{vaswani2017attention} on the WMT14 English-German translation dataset \cite{bojar2014findings}, DeepLabV3+ semantic segmentation model on the Cityscapes dataset \cite{cordts2016cityscapes}, Wav2Vec speech recognition model \cite{baevski2020wav2vec} on the LibriSpeech dataset \cite{panayotov2015librispeech} and many others.

However, there are concerns with this approach. Some argue that we are overfitting on a small set of single or few task benchmarks \cite{raji2021ai,aburass2024quantifying}, leading to solutions that don't generalize as they overfit. On the other hand, others argue that less mainstream fields, such as atmospheric science, struggle to evolve due to the lack of such benchmarks \cite{dueben2022challenges}. A similar issue was identified in the domain of UAVs and aerial image understanding in \cite{marcu2024quantifying}. In the work of \cite{marcu2023self}, they introduce Dronescapes, a dataset for UAVs on three tasks: semantic segmentation, depth estimation and camera normals estimation, which is also a starting point of this work.

Regarding the \textbf{Data (1)} component, recent trends in Machine Learning lean towards massive pre-training followed by task-specific fine-tuning in domains like language modeling (GPT series \cite{radford2018improving}, trained on 40B text tokens), vision transformers (ViT \cite{dosovitskiy2020image}, trained on $\sim$15M images), language-image understanding (CLIP \cite{radford2021learning}, trained on 400M image-text pairs) or segmentation (SAM \cite{kirillov2023segment}, trained on 11M images and 1.1B segmentation masks). These methods provide great results showcasing the need for more high-quality large-scale and multi-modal datasets. Multi-modality refers to the use of multiple types of sensors together to achieve some goals or tasks. For example, combining images with depth information or text descriptions provides a richer understanding than using images alone. While previous works like Dronescapes \cite{marcu2023self} relied on manual curation or specific rigid scripts, this work introduces VRE to automate the dependency management of such modalities.

In this work we introduce an automated data-pipeline for creating multi-modal Machine Learning datasets, called Video Representations Extractor, or VRE for short.

On the \textbf{Models (2)} side, recent trends have been towards large and very large Transformer-based models, with hundreds of millions and billions of parameters. Through techniques like Masked Auto Encoders (MAE), they do large pre-training on generic and easily acquirable data (i.e. RGB only or text only) followed by task-specific fine-tuning (i.e. chatbots or object recognition) \cite{devlin2019bert,he2022masked}. Recent works, such as \cite{bachmann2022multimae}, leverage MAE-based solutions for Multi Task Learning (MTL): depth estimation and semantic segmentation, while \cite{lu2024unified,mizrahi20234m,bai2022ofasys} extend this approach to new modalities (i.e. image, text, audio, action prediction for robotics etc.). These modalities can also be represented as a graph that models the relationships between them \cite{zamir2018taskonomy,leordeanu2021semi,haller2021self,marcu2023self,pirvu2023multi}. Such advances are driven by the rise of foundation models pre-trained on massive datasets, enabling zero-shot prediction via prompting and efficient fine-tuning, as seen in \cite{radford2021learning,kirillov2023segment}.

In this work, we try to close the loop on the relationship between the Data (1) and Model (2) through the means of our modular VRE data-pipeline. This data-pipeline is a Machine Learning specific tool aiming at optimizing the relationship between model design and signal-rich training datasets. As an initial hint of this, using VRE's configuration options, the authors of PHG-MAE \cite{mihai2025probabilistic} have co-designed the modalities for their multi-modal MAE-based algorithm in order to improve the efficiency and effectiveness of their training and model performance for the tasks of semantic segmentation, depth and camera normals estimation.

Regarding \textbf{Actions (3) and predictions}, we can observe that it is a much less explored and researched area. Usually this is enabled by the R\&D on the models side, followed by a deployment procedure. Once a model is deployed it is assumed frozen (most of the time) and it becomes more of an engineering problem to run inference and serve predictions reliably without breaking the existing system. This suggests that the neural network is usually used as a module of a larger system, with this hybrid being referred to as Software 2.0 \cite{software-2-0}, where standard "1.0" procedural code is mixed with neural network predictions to make intelligent actions. One of the main questions and trade-offs is related to where the inference computation happens: on device or on some external server. The first requires the device to have greater compute power, which can increase weight, decrease battery life or increase overall latency. The second solution adds a communication layer between the device and the processing node, which adds variance due to connection issues. Some argue that a distributed system is required to achieve end-to-end real-time performance \cite{becker2020demystifying} with specialized nodes doing specialized tasks, like object recognition. Others propose neural architectural changes to reduce the inference duration variance \cite{liu2022prophet} caused by things like object proposals which can be dynamic based on the input image. Nonetheless, solving these latency-performance trade-offs enables intelligent and autonomous devices, like autonomous vehicles \cite{chen2022level} or drones for use-cases like flood detection \cite{hernandez2022flood}, power line failures \cite{ayoub2021real} or search and rescue assistance \cite{alsamhi2022uav}.

In this paper, we study a simpler use-case: deploying real-time models for semantic segmentation and depth estimation on a laptop consumer GPU (NVIDIA RTX 4050) for local inference, as well as remote inference via a cloud server with a consumer GPU (NVIDIA RTX 2080 Ti). We also study the case of real-time streaming from a phone camera and analyze the trade-offs between the setups. In short, our main contributions presented in this paper are on the data-processing and automation side, as well as model deployment on consumer grade GPUs.

\section{Video Representations Extractor data-pipeline for \\ multi-modal machine learning}

\begin{figure}[H]
    \centering
    \includegraphics[width=1\linewidth]{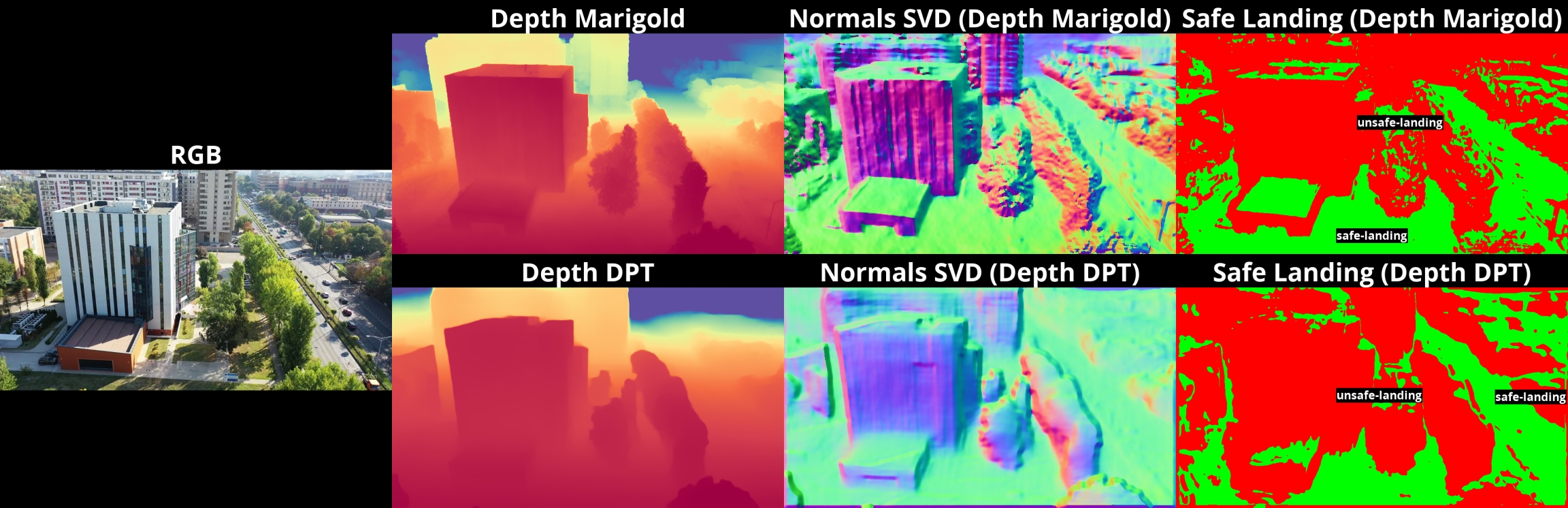}
    \caption{VRE showcase. We present six exported representations on top of the RGB frame. The first two are pre-trained neural networks (DPT \cite{ranftl2021vision} and Marigold \cite{ke2024repurposing}) and depend only on RGB frames. Next, we derive two camera normals representations using a SVD-based algorithm \cite{hartley2003multiple}, which requires the depth maps. Lastly, we derive safe-landing areas by thresholding the camera normals maps like in the newly introduced Dronescapes+VRE \cite{mihai2025probabilistic} dataset.}
    \label{fig:vre-showcase}
\end{figure}

To streamline multi-modal training for video scene understanding, we developed a data-pipeline, named Video Representations Extractor (or VRE for short), which we have also open-sourced \footnote{\url{https://github.com/meehai/video-representations-extractor/}}. We discuss architectural decisions and how it can be used to create new datasets for robotics use-cases beyond UAVs. We also discuss data-pipeline strategies, including multi-gpu batching and real-time streaming. Finally, we present a case study where the authors of \cite{mihai2025probabilistic} extend the Dronescapes dataset \cite{marcu2023self} using VRE.

\subsection{VRE main loop}

A VRE process works on a single accelerator (CPU, GPU etc.), a single video (list of frames) and it executes a single representation at a time. By default, it works on batches, where the batch size is defined globally, with the ability to override this option at representation level. For example, we can process many RGB frames at once, however for learned representations (i.e. neural networks), the batch size must be capped by memory requirements, like the (V)RAM capacity of the accelerator. We use the term 'accelerator' to denote either CPU (regular representations) or GPU (neural representations), with the ability to generalize to other custom accelerators, such as TPUs, NPUs etc. While VRE does not do scheduling at multi-video or multi-representation level, we present later how multiple videos and multiple GPUs can be used to parallelize this simple approach. Algorithm \ref{alg:vre-algorithm} shows the main VRE loop. At its core this is all VRE does, but getting this right is not a trivial task.

\begin{algorithm}
\caption{Video Representations Extractor main loop}\label{alg:cap}
\begin{algorithmic}
    \State $video \gets [frame1, frame2, ..., frameN] $
    \State $representations \gets [RGB, Mask2Former(params), DPT(params), ...]$
    \For{$repr$ in $topo\_sort(representations)$}
        \State $batches \gets make\_batches(video, batch\_size)$ \Comment{batch\_size=1 if streaming}
        \For{$batch\_of\_frames$ in $batches$}
            \If{not $already\_computed(batch\_of\_frames)$}
                \State $out\_repr = repr.compute(batch\_of\_frames, [out\_deps])$
                \State $img\_repr = repr.make\_images(out\_repr)$ \Comment{optional}
            \EndIf
            \State $store\_on\_disk(batch, out\_repr, img\_repr)$ \Comment{only in batch mode}
        \EndFor
        \State $store\_repr\_metadata(batches)$ \Comment{stats about this representation}
    \EndFor
    \State $store\_run\_metadata(video)$ \Comment{stats about this video run}
\end{algorithmic}
\label{alg:vre-algorithm}
\end{algorithm}

The input to the process is a list of frames and the output is the binary-encoded format of each representation, either stored on disk (batch mode) or consumed and dropped (streaming mode). It also produces a process-specific metadata that uniquely identifies the run. Additionally, each representation contains a secondary frame-level metadata file with information about frame computation duration, storage options and a reference to the run metadata. More about the data format and the metadata is discussed in Section \ref{subsec:representation}.

\subsection{Representation}
\label{subsec:representation}

A representation is the basic block of VRE and is roughly equivalent to a modality in broader Machine Learning terms. It has two components:  the definition and the computation. The definition of a representation consists of a unique name, a set of parameters and a list of dependencies on other representations. The computation is the code required to produce the representation the required interface is as simple as:

\begin{lstlisting}
out_repr = repr.compute(batch_of_frames, [out_deps])
img_repr = repr.make_image(out_repr) # optional
\end{lstlisting}

These functions have access to the previously computed representations it depends on, creating a Directed Acyclic Graph (DAG). VRE schedules the running order based on a topological sorting and fails if the dependencies are not solvable, i.e. one or more dependencies are not defined or there are cycles.

Representations are parametrized, which allows us to define multiple ones with different parameters resulting in different computation. An example of this can be seen in Figure \ref{fig:vre-showcase} where the same representation code (SVD-based camera normals from depth), but with slightly different parameters and dependencies (different depth maps as input) produce two distinct camera normals representations. The same concept is repeated to produce two different 'safe landing' semantic maps from two distinct normal maps.

\paragraph{\textbf{Defining the representation}}
The representations that we want to compute are instantiated based on a YAML-based configuration file. In this configuration file we must define the unique name of the representation, its parameters as well as its dependencies (if any). If the dependencies are not properly provided, then $topo\_sort(representations)$ will fail with an appropriate error. The configuration file looks roughly like this:

\begin{lstlisting}
globals:{batch_size:10}
representations:{
  rgb:{type:color/rgb, deps:[],    params:{}},
  hsv:{type:color/hsv, deps:[rgb], params:{batch_size:5}}
}
\end{lstlisting}

\paragraph{\textbf{Running VRE on a video.}} To run VRE on a video (i.e. \textit{video.mp4}) with the above configuration, you need to run the following command:

\begin{lstlisting}
vre video.mp4 -frames 1..N -o out_dir -config_path cfg.yml
\end{lstlisting}

After the VRE process is done, it creates a disk-based data structure:

\begin{lstlisting}[language=c]
| video.mp4
| cfg.yml
| out_dir/
  | .logs/[run_metadata_ID.json, log_ID.txt, ...]
  | rgb/
    | representation_metadata.json
    | npz/[1.npz, ..., N.npz]
  | hsv/
     | representation_metadata.json
    | npz/[1.npz, ..., N.npz]
    | jpg/[1.jpg, ..., N.jpg]
  ... (other representations if defined in cfg.yml) ...
\end{lstlisting}

In batching mode, we want to store the data on the disk such that it can be later loaded for various use-cases, such as training a neural network on the representations-enriched dataset. We discuss batching vs. streaming in Section \ref{subsec:batching-vs-streaming}. This disk-based data structure leverages the rise of fast SSDs, enabling the loading of large batches of data into RAM for efficient usage.

\paragraph{\textbf{VRE metadata.}} For each VRE run, there is a run metadata, with a unique ID, containing information about the list of frames that were processed. Moreover, in each representation, there is a representation metadata file which contains information about each frame, including a reference to the run metadata ID. These metadata files are the basic blocks for state introspection and scheduling. At the start of each VRE run, the tool loads all the metadata files to get the list of all the frames that still need computing and it only schedules the remaining ones, to avoid duplicate work between multiple runs. If VRE is called twice on the same set of frames, the second time it should do no computation. On the other hand, if the first VRE process was stopped halfway, then the second one should one process the remaining ones, effectively implementing a re-entrancy mechanism. Additionally, all the writes to the $representation\_metadata.json$ files are atomic. This enables running multiple VRE processes at the same time on the same subsets of a video. We discuss multi-gpu strategies in Section \ref{subsec:multi-gpu-strategies}.

\paragraph{\textbf{Re-entrancy and disk/memory differences}} Earlier we've hinted that through the metadata information, we have implemented a re-entrancy mechanism to avoid double computation. However, having a representation stored on disk does not always mean it's usable as-is. For example, in the case of semantic segmentation, the information can be stored in three ways: raw predictions as either floating-point logits (1), softmaxed probabilities (2) or argmaxed class indices  as unsigned integers (3). Upon loading the data from the disk, some information may be lost compared to recomputing it from scratch. As this is an application-specific detail, VRE doesn't impose hard limitations. To support this duality (disk vs. memory), a representation must implement the following interface as well, besides $repr.compute()$ defined earlier:

\begin{lstlisting}[language=c]
repr.memory_to_disk(out_repr, path) # store
out_repr = repr.disk_to_memory(path) # load
\end{lstlisting}

\subsection{Data processing strategies: batching vs. streaming}
\label{subsec:batching-vs-streaming}
\begin{figure}[h]
    \centering
    \includegraphics[width=1\linewidth]{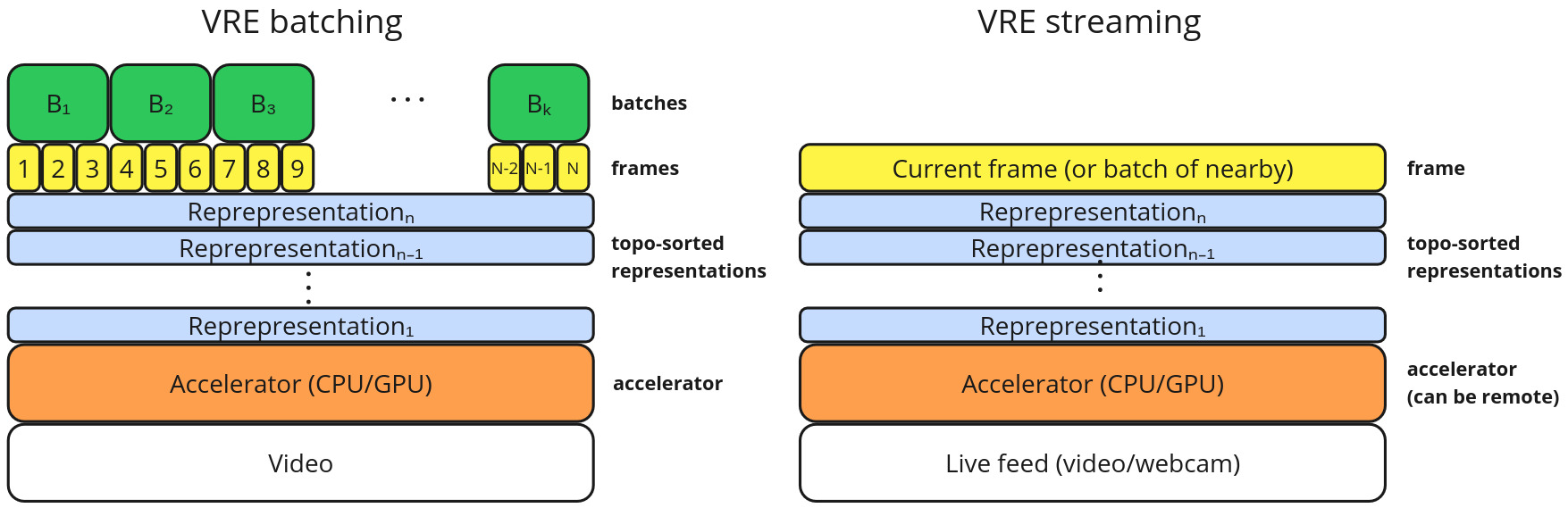}
    \caption{VRE processing strategies. Left: the standard batched strategy. We split the frames in batches and then each batch is processed by the representation's algorithm, followed by a step where the results are stored on the disk. Right: the streaming strategy. In this mode the input is a live video stream (webcam, camera phone etc.) Each frame, or nearby ones (if needed), are processed sequentially by all representations.}
    \label{fig:vre-main-loop}
\end{figure}

Data pipelines face a dichotomy between batching and streaming. The first refers to scheduling large amounts of data in an offline manner for reliable computation and later usage. The second one refers to handling the data stream as it arrives, trading reliability for low latency. At its core, VRE supports both modes, see Figure \ref{fig:vre-main-loop}.

\noindent\textbf{Batching mode.} This is the default mode. We use it to schedule entire videos, compute representations and store them on the disk for later usage. Later usage can refer to Machine Learning tasks, such as training a new neural network using the exported data. Each representation is processed independently, based on the topological sorting earlier described. Ideally, we want to set the the batch size such that we maximize the occupancy of the accelerator (i.e. GPUs). To simplify integration, the exported data can be used as-is using the provided ML-ready data reader implemented in python, though the format is language-agnostic. This mode is aimed at long-running runs and implements various reliability mechanisms, such as retrying with a smaller batch-size in case of OOM errors, a re-entrant mechanism (skipping previously computed frames) and exhaustive diagnostics and log files of each run.

\begin{figure}[h]
    \centering
    \includegraphics[width=0.65\linewidth]{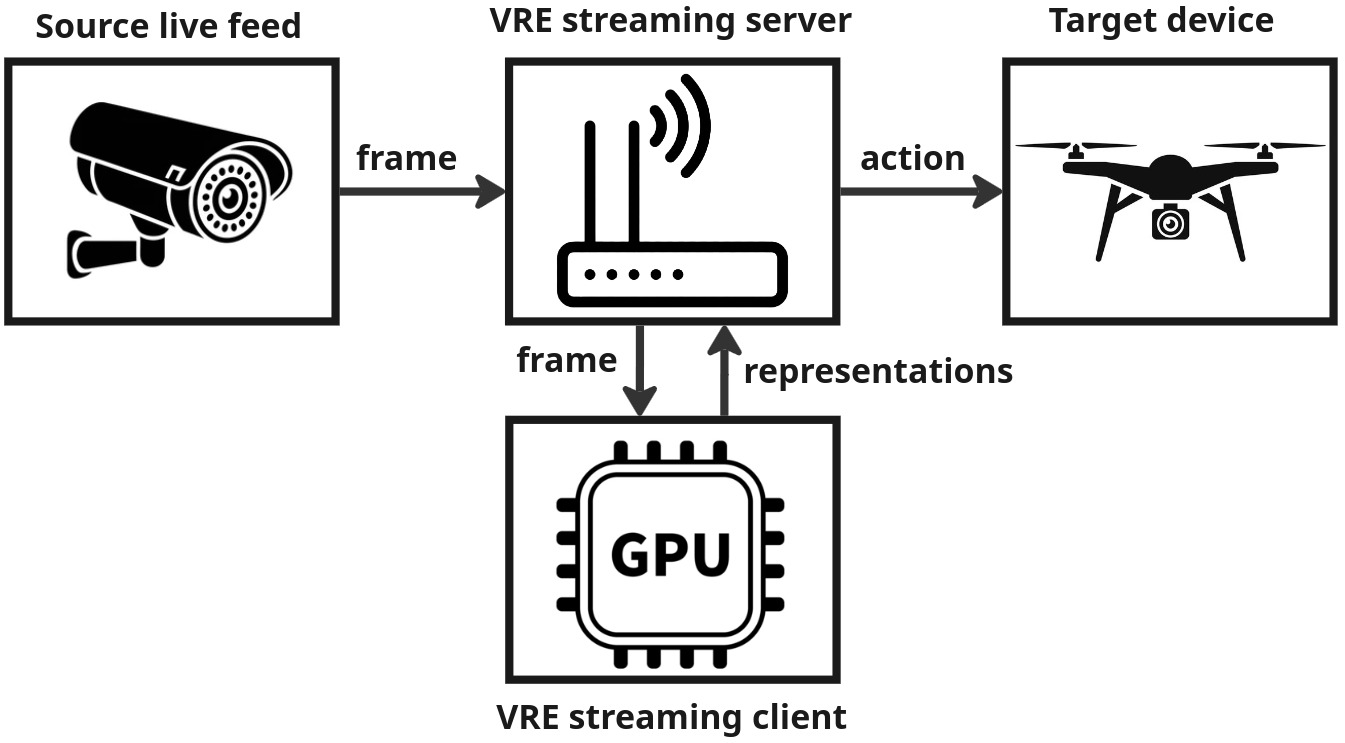}
    \caption{VRE streaming architecture. We read frame by frame from the source (i.e. drone camera) in the VRE streaming server, process it on the VRE streaming client (i.e. cloud or local GPU), analyze the results and pass the actions to the target (i.e. drone controller). Notably, all these components can live on the same machine but they can also communicate through the network.}
    \label{fig:streaming-mode-integration}
\end{figure}

\noindent\textbf{Streaming mode.} VRE also supports streaming mode, where the focus is less on reliability and more on fast inference. In this mode, we disable disk operations and optional features to maximize speed, like computing the human-friendly image representation ($repr.make\_image(out\_repr)$). In Figure \ref{fig:streaming-mode-integration}, we provide a basic diagram of how VRE in streaming mode can be integrated in a larger system. Here, the frames come from an external source (i.e. video, webcam, phone camera etc.), are processed on an external machine (i.e. cloud GPU) and then are used to control an external device (i.e. robot, drone etc.). The experiments section will showcase this case as well.

\subsection{Multi-GPU batching strategies}
\label{subsec:multi-gpu-strategies}

\begin{figure}[h]
    \centering
    \includegraphics[width=1\linewidth]{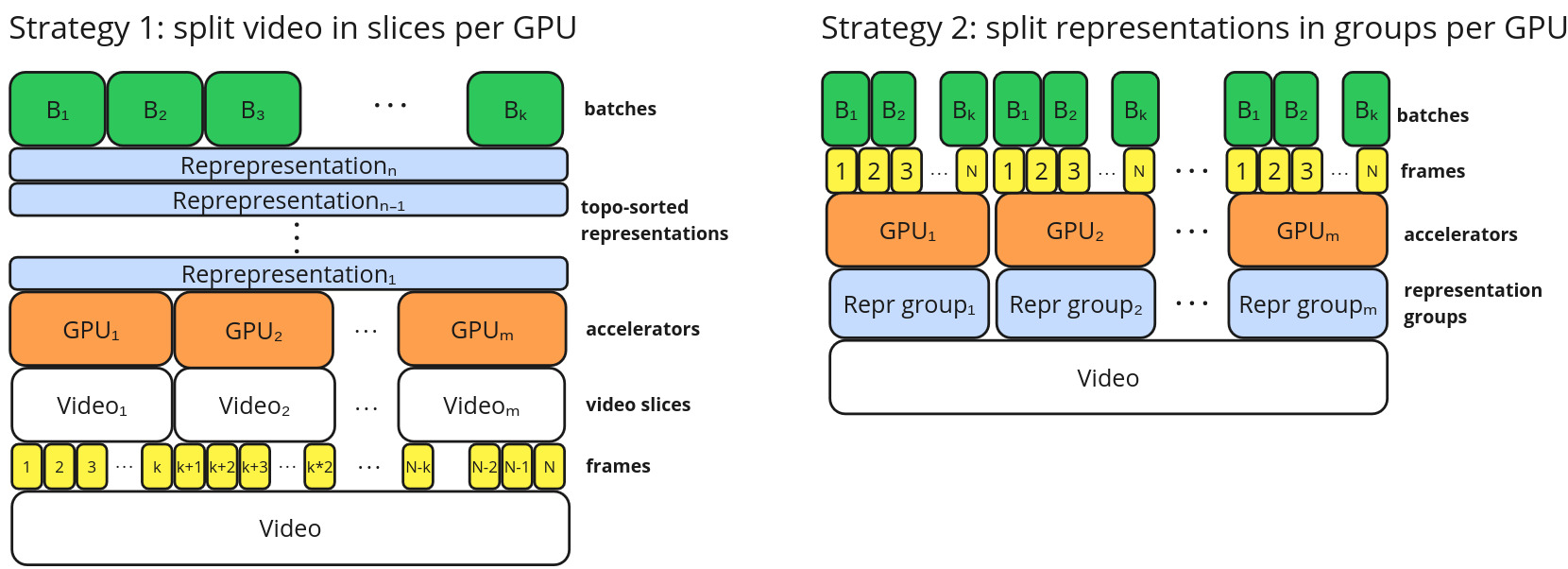}
    \caption{VRE multi-gpu batching strategies. Strategy 1: Slice the video in multiple independent chunks. Strategy 2: split the video's representations in sub-groups.}
    \label{fig:vre-main-loop-multigpu}
\end{figure}

In batching mode, if multiple accelerators are available VRE provides tools to maximize the usage, i.e. nodes with $\ge$ 1 GPUs. In Figure \ref{fig:vre-main-loop-multigpu}, we present two different multi-gpu strategies that can be applied by spawning multiple VRE processes on a single video.

\paragraph{\textbf{Strategy 1}} is the simplest and most effective one. The video is split in multiple independent chunks of frames, each chunk being assigned to one VRE process (and one accelerator/GPU). This strategy ensures consistency and maximizes utilization, as GPUs finish at roughly the same time if the videos are similarly sized. Furthermore, this strategy also works on distributed environments, allowing us to schedule a video on multiple machines as well, given that we don't overlap the chunks. The user must only optimize the configuration for a single GPU  (i.e. tune batch size, parameters etc.), followed by simply applying Strategy 1 which splits the video in chunks across multiple GPUs. VRE natively supports this case via the \textit{-{}-frames A..B} CLI flag. Through this, one can start many VRE processes (one on each GPU) on subsets of the same video. Note that any potential race conditions on the same output directory are solved by using atomic write operations, as we explained in the previous section regarding metadata. Furthermore, we provide a helper CLI tool, \textit{vre\_gpu\_parallel}, which can be used to implement Strategy 1 automatically:

\begin{lstlisting}
CUDA_VISIBLE_DEVICES=0,1,2,3 vre_gpu_parallel \
  VIDEO -o DIR --config_path CONFIG --frames 1..100
Executing: CUDA_VISIBLE_DEVICES=0 vre VIDEO \
  --frames 1..25 -o DIR --config_path CONFIG
...
Executing: CUDA_VISIBLE_DEVICES=3 vre VIDEO \
  --frames 76..100 -o DIR --config_path CONFIG
\end{lstlisting}

\paragraph{\textbf{Strategy 2}} comes as a fine-tuning on top of Strategy 1. In some cases, we may want to compute various representations of the same video on different GPUs as well, especially if they don't depend on each other. Furthermore, we may want to compute non-GPU representations (i.e. not neural networks), like \textit{color/hsv} simply on a CPU without other representations waiting for it while keeping the GPU idle. To do so, we define representation groups based on the topological sort where each group is independent and has no dependencies outside of it. For example, in Figure \ref{fig:vre-showcase}, the top row and the bottom row are independent groups which can be computed in parallel. The CLI also provides a way to select only a subset of representations from a configuration file, so each GPU only processes its representation group.

In conclusion, the two strategies are complement each other and can improve the efficiency of the overall process if multiple accelerators are available.

\subsection{VRE repository} At the time of writing, the following algorithms and pre-trained modalities are implemented and ready to use.
\label{subsec:vre-repository}

\begin{itemize}
    \item Color: RGB, HSV
    \item Edges: Canny \cite{canny1986computational}, DexiNed \cite{poma2020dense}
    \item Optical flow: RAFT \cite{teed2020raft}, RIFE \cite{huang2022real}
    \item Depth estimation - DPT \cite{ranftl2021vision}, Marigold \cite{ke2024repurposing}
    \item Normal maps: SVD (from depth) \cite{hartley2003multiple}
    \item "Soft" segmentation: FastSAM \cite{zhao2023fast}, Generalized Boundaries \cite{leordeanu2014generalized}, Halftone
    \item Semantic segmentation: Mask2Former \cite{cheng2022masked}, PHG-MAE-Distil \cite{mihai2025probabilistic}
    \item Objects detection: YOLO \cite{tian2025yolov12}
\end{itemize}

Upon representation instantiation, the weights of these representations (for learned representations only) are downloaded locally from a cloud storage. Implementing a new representation is as simple as implementing a shared interface defining a few methods, such as \textit{compute(batch\_of\_frames, dependencies)} and \textit{make\_image(frame, computed\_result)}. For \textit{learned representations} (i.e. neural networks), one must also implement two more methods: \textit{load\_weights(path)} and \textit{unload\_weights()} which are used to load the networks and clear the memory during execution. Moreover, one can implement more fine-grained controls, such as the previously mentioned $disk\_to\_memory$ and $memory\_to\_disk$ functions.

\section{Experiments}

This section showcases the capabilities of the Video Representations Extractor (VRE) data-pipeline on relevant Machine Learning workloads. We first look at the batched case, as introduced in Section \ref{subsec:batching-vs-streaming}, providing three simple-to-complex experiments. Then, we analyze real-time ML streaming by presenting two experiments (local GPU and remote GPU offloading) with two models implemented in the VRE repository for semantic segmentation and depth estimation. All the batched experiments are evaluated based on the reported duration in the metadata files. Furthermore, for the streaming experiments, we report the frames per second (FPS). The experiments are run on one of the 3 available accelerators: \textit{Intel Xeon Gold 5218} CPU, one to eight \textit{NVIDIA RTX 2080 Ti} GPUs or one laptop \textit{NVIDIA RTX 4050} GPU.

\subsection{Batched export: Simple RGB and HSV only}

\begin{figure}[h]
    \centering
    \includegraphics[width=0.8\linewidth]{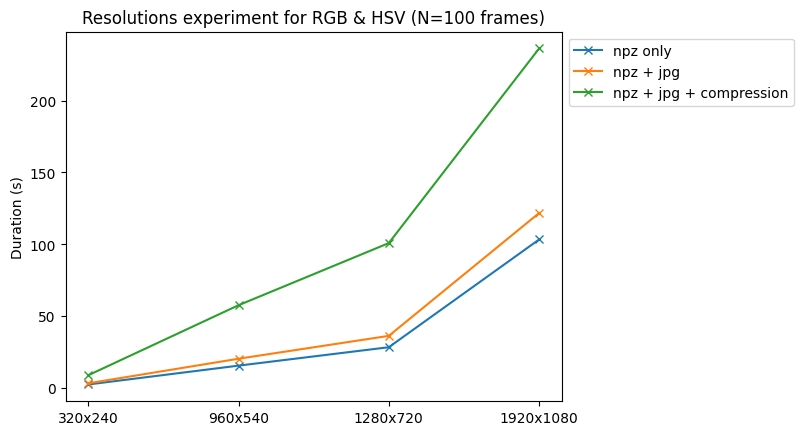}
    \caption{Simple export experiment. We compare three output formats for two computed representations: RGB and HSV.}
    \label{fig:experiment-simple-export}
\end{figure}

We begin with a baseline experiment evaluating the extraction of RGB and HSV representations. RGB extraction involves direct frame copying, while HSV conversion represents a standard pixel-wise transformation that does not require hardware acceleration. We evaluate four video resolutions (240p, 540p, 720p, 1080p) across three export configurations: binary only (.npz), binary plus images (.npz + .jpg), and compressed binary plus images. For consistency, we process a fixed sequence of 100 frames for all variations. These experiments demonstrate VRE's performance in large-scale batch processing, with results detailed in Figure \ref{fig:experiment-simple-export}.

As expected, processing latency correlates with storage complexity (compressed vs. uncompressed), frame resolution, and I/O settings. Notably, enabling compression yields a significant storage reduction of approximately $2.6\times$, decreasing from 2.4GB to 907MB for 1080p HSV export. This comes at a computational cost, increasing processing time by a factor of 1.93, from 121.8s to 236.2s. This highlights a configurable space-time trade-off essential for managing large-scale video datasets. Since these specific operations are CPU-bound, batch size has negligible impact on performance.

\subsection{Batched export: RGB, PHG-MAE-Distil and DPT}

\begin{figure}[h]
    \centering
    \includegraphics[width=1\linewidth]{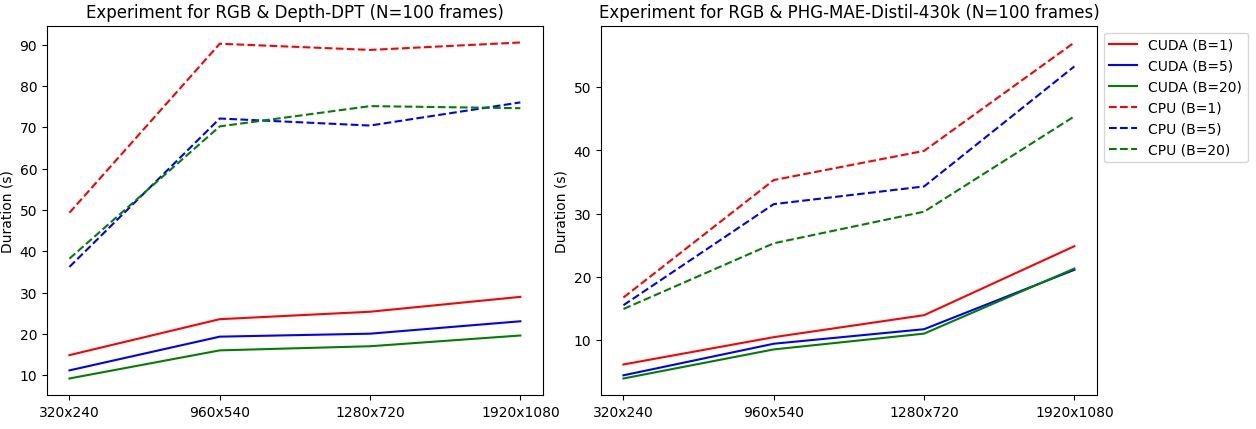}
    \caption{Batched export with neural networks. Left: DPT \cite{ranftl2021vision}. Right: PHG-MAE-Distil \cite{mihai2025probabilistic}. We vary batch size, accelerator and video resolution.}
    \label{fig:experiments-batched-export}
\end{figure}

In this experiment, we want to work with GPU-runnable representations. We analyze the time it takes to compute two neural networks by testing four different video resolutions, three different batch sizes as well as computing them on a CPU vs. a GPU. The results can be seen in Figure \ref{fig:experiments-batched-export}.

First, we observe that the GPU (CUDA) variant consistently outperforms the CPU one on each experiment, regardless of batch size. This is expected as ML models are optimized for parallel GPU execution. For the DPT model we observe about a 5x improvement, while for the PHG-MAE-Distil model, we see a 2.5-3x improvement depending on batch size. Moreover, we observe a constant improvement as we increase the batch size on both CPU and GPU for all the video resolutions. For the DPT model we see about a 1.5x improvement between the B=1 and B=20. This holds even for the PHG-MAE-Distil model, which sees a 1.1-1.3x speed-up. While these improvements may not be huge, we should always aim at maximizing the usage of our accelerators as each image in batch-mode is independent from each other, allowing for parallel processing. The only reason we would use a lower batch size is due to memory constraints.

\subsection{Batched export: Dronescapes config - 17 representations}
\label{subsec:complex-batch-export}

\begin{figure}[h]
    \centering
    \includegraphics[width=1\linewidth]{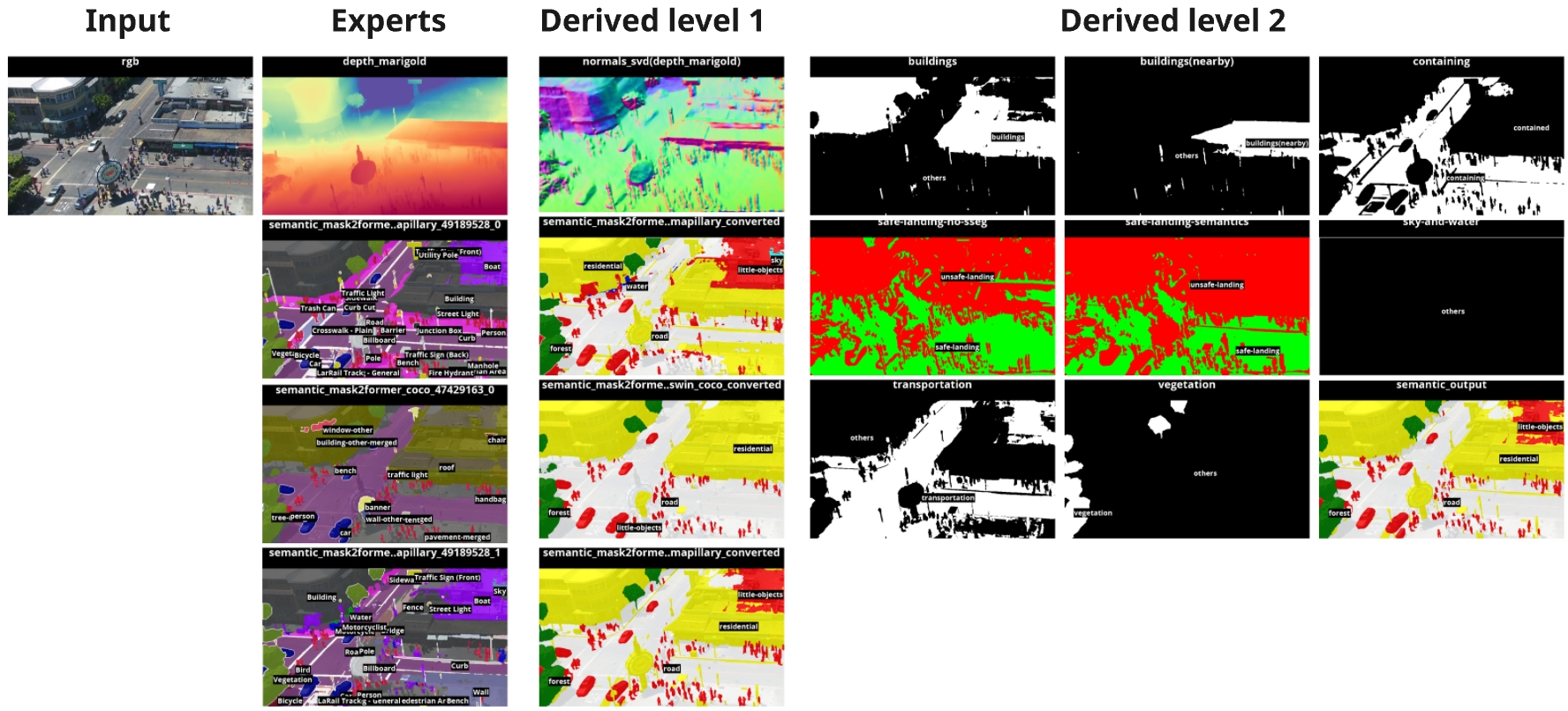}
    \caption{Dronescapes+VRE: All the extracted neural and derived intermediate representations. All are generated starting only from the RGB image.}
    \label{fig:results-qualitative-dronescapes2}
\end{figure}

We analyze the work of \cite{mihai2025probabilistic} which has used our tool to expand the original Dronescapes dataset \cite{marcu2023self} from $\sim$23K training frames to about $\sim$148K. As they've released the VRE config file, we also it for analysis on 100 frames of a video at the same resolution (960x540). In Figure \ref{fig:results-qualitative-dronescapes2} we provide a qualitative sample of the exported representations (or modalities in ML terms).

\begin{figure}[h]
    \centering
    \includegraphics[width=1\linewidth]{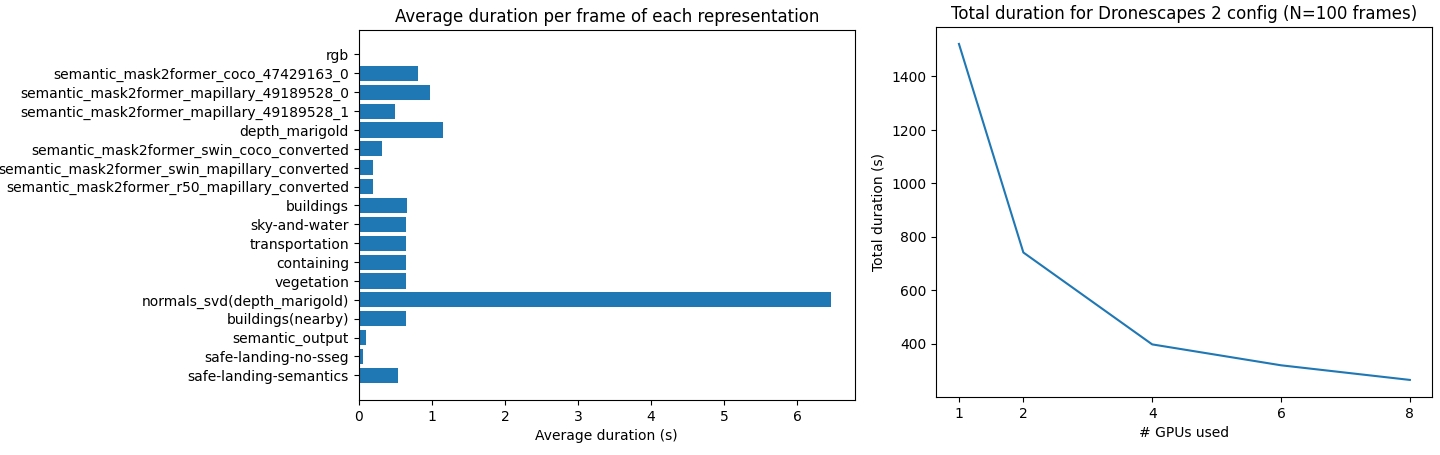}
    \caption{Results VRE on the Dronescapes config on RTX 2080 Ti GPUs. Left: bar plot with the average duration of each representation per frame on a single GPU. Right: Multi-GPU scaling. }
    \label{fig:results-dronescapes2}
\end{figure}

We observe the three-level nesting of the VRE process. The neural networks (labeled 'Experts') require only an RGB frame as input. Then, the first derived intermediate modalities are the camera normals from depth and the semantic segmentation mapping: from the original datasets of Mask2Former to the Dronescapes+VRE set of classes. The final level of derived modalities are built on top of the previous two levels, which are already available at that point due to topological sorting. In Figure \ref{fig:results-dronescapes2}, we present quantitative results on duration and multi-gpu scaling.

The left bar plot computes the duration to produce each of the modalities of Figure \ref{fig:results-qualitative-dronescapes2} on a single GPU. We observe that most of the time is spent on a single representation, namely the normals from SVD algorithm \cite{hartley2003multiple}, taking an average of 6 seconds per frame to compute. This is expected, as this algorithm is not easily parallelizable. On the other hand, even neural network representations, such as Mask2Former \cite{cheng2022masked} or Marigold \cite{ke2024repurposing} take about 1 second on average for each frame. In total, it takes about 15.2s to process one frame on one GPU.

On the right side of the figure, we run the same configuration in a multi-gpu setup using Strategy 1 (Section \ref{subsec:multi-gpu-strategies}). We observe that the average computation time drops almost linearly with the number of GPUs when using 2 or 4 GPUs, but then it plateaus. For 8 GPUs it takes about 264 seconds, a drop from 1521 seconds across 100 frames, reaching a 72\% scaling efficiency compared to a theoretical perfect scaling of 190 seconds. This sub-linear scaling likely results from the fact that other resources (such as I/O, RAM or CPU) are also bottlenecking the parallelism.

\paragraph{\textbf{VRE vs. human annotations.}} We compare the cost of generating this data against traditional methods. Fine-grained pixel-level semantic annotation is estimated to require approximately 7 (coarse) to 90 minutes (fine) per frame \cite{das2023urban}, while annotating regression-based frames, such as depth or camera normals, is practically infeasible without specialized software or hardware (i.e. LiDAR). In contrast, VRE eliminates the need for human effort during generation, shifting the cost to computation. The pipeline processes a single frame across all modalities in $\sim$15.2s on a single GPU, dropping to an average of 2.6s when using 8 GPUs. This yields a throughput increase of 30-350$\times$ compared to coarse/fine manual labeling on a single GPU (and significantly higher with multi-GPU scaling). While a direct quality comparison is challenging given that human annotation remains the gold standard, algorithmic strategies can be applied to mitigate the gap, like employing ensemble methods, such as taking the median of many segmentation models at once.

\subsection{Streaming: Real-time inference on a consumer-grade GPU}

\begin{figure}[h]
    \centering
    \includegraphics[width=0.7\linewidth]{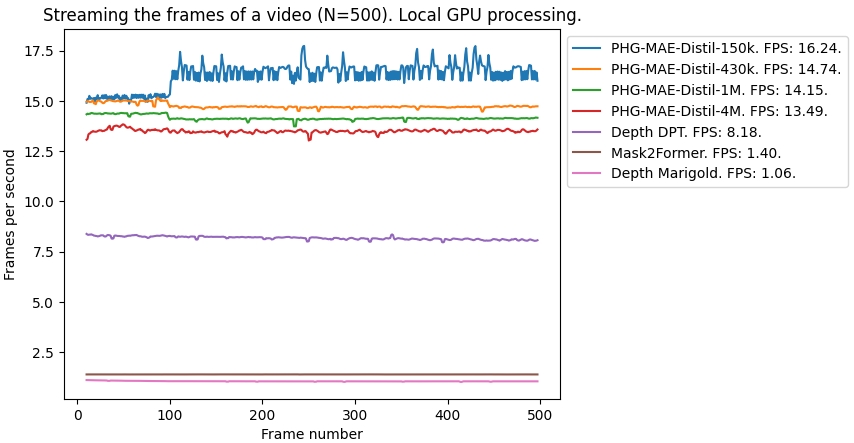}
    \caption{Streaming the frames of a video through various ML models. Processed on a local GPU.}
    \label{fig:experiment-streaming-local}
\end{figure}

In this experiment we test the streaming capabilities of VRE against various ML models, following the architecture presented in Figure \ref{fig:streaming-mode-integration}. In this case all the components are local: the source is a video file at a 960x540 resolution, the GPU is local and the output is displayed in a video player. We present the results in Figure \ref{fig:experiment-streaming-local}.

We measure the time spent processing in the VRE tool with no model as well as processing through various models: PHG-MAE-Distil \cite{mihai2025probabilistic} (150k$\sim$4.4M params), Mask2Former \cite{cheng2022masked} Depth DPT \cite{ranftl2021vision} and Depth Marigold \cite{ke2024repurposing}. We observe that the models have quite a low variance during execution, which makes the streaming of ML models quite reliable. Notably, the PHG-MAE-Distil variants can be used for real-time segmentation, while the Depth DPT can be used for real-time depth estimation, which can enable various robotics applications, such as safe navigation through a natural environment. The other two models, while they output higher-quality results, are better suited for batched export achieving less than 2 FPS.

\subsection{Streaming: Real-time ML with a handheld device and remote vs. local processing}

\begin{figure}[h]
    \centering
    \includegraphics[width=0.83\linewidth]{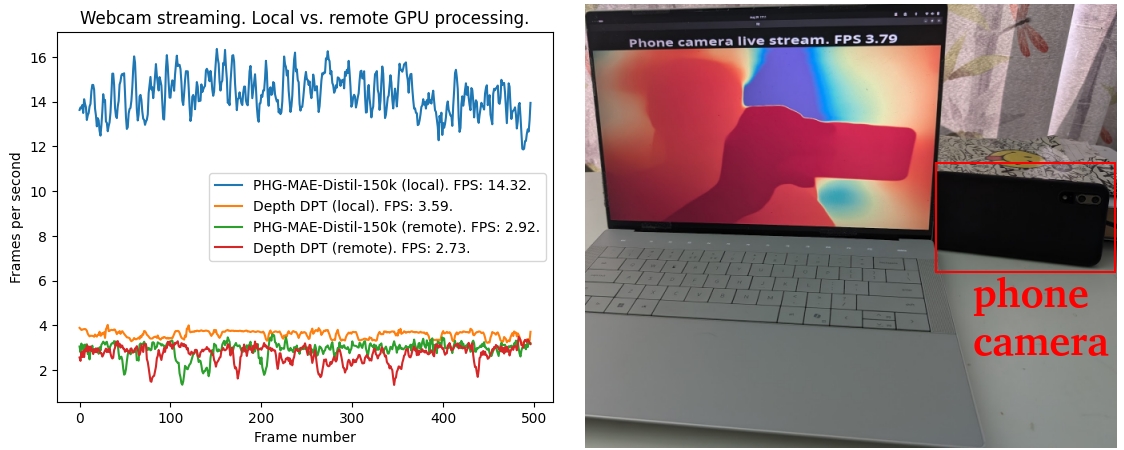}
    \caption{Streaming the frames of a phone camera through various ML models. Processed on a local GPU and on a remote GPU. Left: the FPS results. Right: The live-streaming setup.}
    \label{fig:experiment-streaming-remote}
\end{figure}

In this experiment, we want to analyze the trade-offs introduced by network latency. The setup is as follows: we capture the camera feed from the mobile phone (1) which we then relay through the VRE server (2) to the processing GPU (local or remote) (3). Finally, the processed images are displayed back on the target device, that is the video player on the laptop (4). This maps the same setup as Figure \ref{fig:streaming-mode-integration}. The remote machine is the same as the one used in all the experiments before, while the local machine is the laptop in the image, with a consumer-grade GPU (NVIDIA RTX 4050). We present the results in Figure \ref{fig:experiment-streaming-remote}.

We showcase four results: 2 models and 2 GPUs (local laptop and remote). The blue and orange lines are comparable to the ones reported in the previous experiment (Figure \ref{fig:experiment-streaming-local}), with the drop in FPS being explained by using a laptop GPU instead of a server GPU. On the other hand, the green and red lines present the case where the processing is done remotely. In this case, we see that network latency limits both models to about 2-3 FPS. The main conclusion to be drawn here is that real-time processing is very hard to achieve over a network, thus local computation should be aimed for if possible.

\section{Conclusions}

We introduce a Machine Learning infrastructure data-pipeline aimed at streamlining the creation of multi-modal datasets for training deep neural networks. We present the architectural design and the batched vs. streaming duality, which the tool supports natively. For the batched case, we provide multi-gpu strategies, such as splitting a video in multiple slices or targeting different GPUs with representation groups. We open source the tool alongside a repository of already implemented representations. We then present a case study for how Dronescapes, an aerial image understanding dataset, was created using this tool. We provide experiments for both the batched case, as well as a real-time and near-real-time semantic segmentation and depth estimation streaming pipeline using a handheld phone's camera as live feed.

As future work, our data-pipeline can be improved to support other native video streaming protocols, such as RTP to improve latency. Moreover, the tool works natively on a single node, allowing node-level parallelism, such as multi-gpu setups. However, this approach could be extended to support distributed systems, allowing for a more seamless vertical scaling where nodes can be created and deleted on demand. On VRE repository side, we implement a few off-the-shelf models (Section \ref{subsec:vre-repository}), however additional ones can be integrated, like object tracking, keypoint extraction or video action recognition. Finally, on the dataset creation side, VRE focuses on generating new modalities from raw data, pre-trained models and procedural algorithms directly from videos. However, there are other issues that VRE does not tackle, like raw dataset acquisition (e.g. from YouTube) or dataset curation.

\paragraph{\textbf{Acknowledgements.}} This work is supported in part by projects “Romanian Hub for Artificial Intelligence - HRIA”, Smart Growth, Digitization and Financial Instruments Program, 2021-2027 (MySMIS no. 334906) and "European Lighthouse of AI for Sustainability - ELIAS", Horizon Europe program (Grant No. 101120237).


\begin{thebibliography}{99}
\setlength{\baselineskip}{.45cm}

\bibitem{aburass2024quantifying} {\it Sanad Aburass, Maha Abu Rumman}. Quantifying overfitting: introducing the overfitting index, in {\it 2024 International Conference on Electrical, Computer and Energy Technologies (ICECET)}, IEEE, 1-7, 2024.

\bibitem{alsamhi2022uav} {\it Saeed Hamood Alsamhi, Alexey V. Shvetsov, Santosh Kumar, Svetlana V. Shvetsova, Mohammed A. Alhartomi, Ammar Hawbani, Navin Singh Rajput, Sumit Srivastava, Abdu Saif, Vincent Omollo Nyangaresi}. UAV computing-assisted search and rescue mission framework for disaster and harsh environment mitigation, {\it Drones}, 6(7):154, 2022.

\bibitem{ayoub2021real} {\it Naeem Ayoub, Peter Schneider-Kamp}. Real-time on-board deep learning fault detection for autonomous UAV inspections, {\it Electronics}, 10(9):1091, 2021.

\bibitem{bachmann2022multimae} {\it Roman Bachmann, David Mizrahi, Andrei Atanov, Amir Zamir}. MultiMAE: Multi-modal multi-task masked autoencoders, in {\it European Conference on Computer Vision}, Springer, 348-367, 2022.

\bibitem{baevski2020wav2vec} {\it Alexei Baevski, Yuhao Zhou, Abdelrahman Mohamed, Michael Auli}. wav2vec 2.0: A framework for self-supervised learning of speech representations, {\it Advances in Neural Information Processing Systems}, 33:12449-12460, 2020.

\bibitem{bai2022ofasys} {\it Jinze Bai, Rui Men, Hao Yang, Xuancheng Ren, Kai Dang, Yichang Zhang, Xiaohuan Zhou, Peng Wang, Sinan Tan, An Yang, et al.}. OFASys: A multi-modal multi-task learning system for building generalist models, {\it arXiv preprint arXiv:2212.04408}, 2022.

\bibitem{becker2020demystifying} {\it Pedro H.E. Becker, Jose Maria Arnau, Antonio Gonz\'alez}. Demystifying power and performance bottlenecks in autonomous driving systems, in {\it 2020 IEEE International Symposium on Workload Characterization (IISWC)}, IEEE, 205-215, 2020.

\bibitem{bojar2014findings} {\it Ond\v{r}ej Bojar, Christian Buck, Christian Federmann, Barry Haddow, Philipp Koehn, Johannes Leveling, Christof Monz, Pavel Pecina, Matt Post, Herve Saint-Amand, et al.}. Findings of the 2014 workshop on statistical machine translation, in {\it Proceedings of the Ninth Workshop on Statistical Machine Translation}, 12-58, 2014.

\bibitem{canny1986computational} {\it John Canny}. A computational approach to edge detection, {\it IEEE Transactions on Pattern Analysis and Machine Intelligence}, PAMI-8(6):679-698, 1986.

\bibitem{chen2022level} {\it Li Chen, Tutian Tang, Zhitian Cai, Yang Li, Penghao Wu, Hongyang Li, Jianping Shi, Junchi Yan, Yu Qiao}. Level 2 autonomous driving on a single device: Diving into the devils of OpenPilot, {\it arXiv preprint arXiv:2206.08176}, 2022.

\bibitem{cheng2022masked} {\it Bowen Cheng, Ishan Misra, Alexander G. Schwing, Alexander Kirillov, Rohit Girdhar}. Masked-attention mask transformer for universal image segmentation, in {\it Proceedings of the IEEE/CVF Conference on Computer Vision and Pattern Recognition}, 1290-1299, 2022.

\bibitem{cordts2016cityscapes} {\it Marius Cordts, Mohamed Omran, Sebastian Ramos, Timo Rehfeld, Markus Enzweiler, Rodrigo Benenson, Uwe Franke, Stefan Roth, Bernt Schiele}. The Cityscapes dataset for semantic urban scene understanding, in {\it Proceedings of the IEEE Conference on Computer Vision and Pattern Recognition}, 3213-3223, 2016.

\bibitem{das2023urban} {\it Anurag Das, Yongqin Xian, Yang He, Zeynep Akata, Bernt Schiele}. Urban scene semantic segmentation with low-cost coarse annotation, in {\it Proceedings of the IEEE/CVF Winter Conference on Applications of Computer Vision}, 5978-5987, 2023.

\bibitem{devlin2019bert} {\it Jacob Devlin, Ming-Wei Chang, Kenton Lee, Kristina Toutanova}. BERT: Pre-training of deep bidirectional transformers for language understanding, in {\it Proceedings of the 2019 Conference of the North American Chapter of the Association for Computational Linguistics: Human Language Technologies, Volume 1}, 4171-4186, 2019.

\bibitem{dosovitskiy2020image} {\it Alexey Dosovitskiy, Lucas Beyer, Alexander Kolesnikov, Dirk Weissenborn, Xiaohua Zhai, Thomas Unterthiner, Mostafa Dehghani, Matthias Minderer, Georg Heigold, Sylvain Gelly, et al.}. An image is worth 16x16 words: Transformers for image recognition at scale, {\it arXiv preprint arXiv:2010.11929}, 2020.

\bibitem{dueben2022challenges} {\it Peter D. Dueben, Martin G. Schultz, Matthew Chantry, David John Gagne, David Matthew Hall, Amy McGovern}. Challenges and benchmark datasets for machine learning in the atmospheric sciences: Definition, status, and outlook, {\it Artificial Intelligence for the Earth Systems}, 1(3):e210002, 2022.

\bibitem{haller2021self} {\it Emanuela Haller, Elena Burceanu, Marius Leordeanu}. Self-supervised learning in multi-task graphs through iterative consensus shift, {\it arXiv preprint arXiv:2103.14417}, 2021.

\bibitem{hartley2003multiple} {\it Richard Hartley, Andrew Zisserman}. Multiple view geometry in computer vision, Cambridge University Press, 2003.

\bibitem{he2022masked} {\it Kaiming He, Xinlei Chen, Saining Xie, Yanghao Li, Piotr Doll\'ar, Ross Girshick}. Masked autoencoders are scalable vision learners, in {\it Proceedings of the IEEE/CVF Conference on Computer Vision and Pattern Recognition}, 16000-16009, 2022.

\bibitem{hernandez2022flood} {\it Daniel Hern\'andez, Jos\'e M. Cecilia, Juan-Carlos Cano, Carlos T. Calafate}. Flood detection using real-time image segmentation from unmanned aerial vehicles on edge-computing platform, {\it Remote Sensing}, 14(1):223, 2022.

\bibitem{huang2022real} {\it Zhewei Huang, Tianyuan Zhang, Wen Heng, Boxin Shi, Shuchang Zhou}. Real-time intermediate flow estimation for video frame interpolation, in {\it European Conference on Computer Vision}, Springer, 624-642, 2022.

\bibitem{software-2-0} {\it Andrej Karpathy}. Software 2.0, \url{https://web.archive.org/web/20250323195948/https://karpathy.medium.com/software-2-0-a64152b37c35}, 2025. [Online; accessed 04-April-2025].

\bibitem{ke2024repurposing} {\it Bingxin Ke, Anton Obukhov, Shengyu Huang, Nando Metzger, Rodrigo Caye Daudt, Konrad Schindler}. Repurposing diffusion-based image generators for monocular depth estimation, in {\it Proceedings of the IEEE/CVF Conference on Computer Vision and Pattern Recognition}, 9492-9502, 2024.

\bibitem{kirillov2023segment} {\it Alexander Kirillov, Eric Mintun, Nikhila Ravi, Hanzi Mao, Chloe Rolland, Laura Gustafson, Tete Xiao, Spencer Whitehead, Alexander C. Berg, Wan-Yen Lo, et al.}. Segment anything, in {\it Proceedings of the IEEE/CVF International Conference on Computer Vision}, 4015-4026, 2023.

\bibitem{krizhevsky2012imagenet} {\it Alex Krizhevsky, Ilya Sutskever, Geoffrey E. Hinton}. ImageNet classification with deep convolutional neural networks, {\it Advances in Neural Information Processing Systems}, 25, 2012.

\bibitem{leordeanu2021semi} {\it Marius Leordeanu, Mihai Cristian P\^irvu, Dragos Costea, Alina E. Marcu, Emil Slusanschi, Rahul Sukthankar}. Semi-supervised learning for multi-task scene understanding by neural graph consensus, in {\it Proceedings of the AAAI Conference on Artificial Intelligence}, 35:1882-1892, 2021.

\bibitem{leordeanu2014generalized} {\it Marius Leordeanu, Rahul Sukthankar, Cristian Sminchisescu}. Generalized boundaries from multiple image interpretations, {\it IEEE Transactions on Pattern Analysis and Machine Intelligence}, 36(7):1312-1324, 2014.

\bibitem{liu2022prophet} {\it Liangkai Liu, Zheng Dong, Yanzhi Wang, Weisong Shi}. Prophet: Realizing a predictable real-time perception pipeline for autonomous vehicles, in {\it 2022 IEEE Real-Time Systems Symposium (RTSS)}, IEEE, 305-317, 2022.

\bibitem{lu2024unified} {\it Jiasen Lu, Christopher Clark, Sangho Lee, Zichen Zhang, Savya Khosla, Ryan Marten, Derek Hoiem, Aniruddha Kembhavi}. Unified-IO 2: Scaling autoregressive multimodal models with vision language audio and action, in {\it Proceedings of the IEEE/CVF Conference on Computer Vision and Pattern Recognition}, 26439-26455, 2024.

\bibitem{marcu2024quantifying} {\it Alina Marcu}. Quantifying the synthetic and real domain gap in aerial scene understanding, {\it arXiv preprint arXiv:2411.19913}, 2024.

\bibitem{marcu2023self} {\it Alina Marcu, Mihai Pirvu, Dragos Costea, Emanuela Haller, Emil Slusanschi, Ahmed Nabil Belbachir, Rahul Sukthankar, Marius Leordeanu}. Self-supervised hypergraphs for learning multiple world interpretations, in {\it Proceedings of the IEEE/CVF International Conference on Computer Vision}, 983-992, 2023.

\bibitem{mihai2025probabilistic} {\it P\^irvu Mihai-Cristian, Marius Leordeanu}. Probabilistic hyper-graphs using multiple randomly masked autoencoders for semi-supervised multi-modal multi-task learning, {\it arXiv preprint arXiv:2510.10068}, 2025.

\bibitem{mizrahi20234m} {\it David Mizrahi, Roman Bachmann, Oguzhan Kar, Teresa Yeo, Mingfei Gao, Afshin Dehghan, Amir Zamir}. 4M: Massively multimodal masked modeling, {\it Advances in Neural Information Processing Systems}, 36:58363-58408, 2023.

\bibitem{panayotov2015librispeech} {\it Vassil Panayotov, Guoguo Chen, Daniel Povey, Sanjeev Khudanpur}. LibriSpeech: An ASR corpus based on public domain audio books, in {\it 2015 IEEE International Conference on Acoustics, Speech and Signal Processing (ICASSP)}, IEEE, 5206-5210, 2015.

\bibitem{pirvu2023multi} {\it Mihai Pirvu, Alina Marcu, Maria Alexandra Dobrescu, Ahmed Nabil Belbachir, Marius Leordeanu}. Multi-task hypergraphs for semi-supervised learning using earth observations, in {\it Proceedings of the IEEE/CVF International Conference on Computer Vision}, 3404-3414, 2023.

\bibitem{poma2020dense} {\it Xavier Soria Poma, Edgar Riba, Angel Sappa}. Dense extreme inception network: Towards a robust CNN model for edge detection, in {\it Proceedings of the IEEE/CVF Winter Conference on Applications of Computer Vision}, 1923-1932, 2020.

\bibitem{radford2021learning} {\it Alec Radford, Jong Wook Kim, Chris Hallacy, Aditya Ramesh, Gabriel Goh, Sandhini Agarwal, Girish Sastry, Amanda Askell, Pamela Mishkin, Jack Clark, et al.}. Learning transferable visual models from natural language supervision, in {\it International Conference on Machine Learning}, PMLR, 8748-8763, 2021.

\bibitem{radford2018improving} {\it Alec Radford, Karthik Narasimhan, Tim Salimans, Ilya Sutskever, et al.}. Improving language understanding by generative pre-training, {\it OpenAI preprint}, 2019.

\bibitem{raji2021ai} {\it Inioluwa Deborah Raji, Emily M. Bender, Amandalynne Paullada, Emily Denton, Alex Hanna}. AI and the everything in the whole wide world benchmark, {\it arXiv preprint arXiv:2111.15366}, 2021.

\bibitem{ranftl2021vision} {\it Ren\'e Ranftl, Alexey Bochkovskiy, Vladlen Koltun}. Vision transformers for dense prediction, in {\it Proceedings of the IEEE/CVF International Conference on Computer Vision}, 12179-12188, 2021.

\bibitem{teed2020raft} {\it Zachary Teed, Jia Deng}. RAFT: Recurrent all-pairs field transforms for optical flow, in {\it Computer Vision -- ECCV 2020: 16th European Conference, Glasgow, UK, August 23-28, 2020, Proceedings, Part II 16}, Springer, 402-419, 2020.

\bibitem{tian2025yolov12} {\it Yunjie Tian, Qixiang Ye, David Doermann}. YOLOv12: Attention-centric real-time object detectors, {\it arXiv preprint arXiv:2502.12524}, 2025.

\bibitem{vaswani2017attention} {\it Ashish Vaswani, Noam Shazeer, Niki Parmar, Jakob Uszkoreit, Llion Jones, Aidan N. Gomez, {\L}ukasz Kaiser, Illia Polosukhin}. Attention is all you need, {\it Advances in Neural Information Processing Systems}, 30, 2017.

\bibitem{zamir2018taskonomy} {\it Amir R. Zamir, Alexander Sax, William Shen, Leonidas J. Guibas, Jitendra Malik, Silvio Savarese}. Taskonomy: Disentangling task transfer learning, in {\it Proceedings of the IEEE Conference on Computer Vision and Pattern Recognition}, 3712-3722, 2018.

\bibitem{zhao2023fast} {\it Xu Zhao, Wenchao Ding, Yongqi An, Yinglong Du, Tao Yu, Min Li, Ming Tang, Jinqiao Wang}. Fast segment anything, {\it arXiv preprint arXiv:2306.12156}, 2023.

\end{thebibliography}


\end{document}